%% file: main.tex
\documentclass[10pt,twocolumn,letterpaper]{article}

\usepackage[pagenumbers]{cvpr} 
\input{preamble}

\definecolor{cvprblue}{rgb}{0.21,0.49,0.74}
\usepackage[pagebackref,breaklinks,colorlinks,citecolor=cvprblue]{hyperref}

\def\paperID{13349} 
\def\confName{CVPR}
\def\confYear{2024}

\title{Adaptive Bidirectional Displacement for Semi-Supervised Medical Image Segmentation}

\author{Hanyang Chi$^{1}$ \hspace{12pt} Jian Pang$^{1}$ \hspace{12pt} Bingfeng Zhang$^{1}$\thanks{Corresponding author.} \hspace{12pt} Weifeng Liu$^{1}$\\
		$^1$China University of Petroleum (East China)\\ 
		{\tt\small \{chihanyang, jianpang\}@s.upc.edu.cn, \{bingfeng.zhang, liuwf\}@upc.edu.cn}
}

\begin{document}
\maketitle
\input{sec/0_abstract}    
\input{sec/1_intro}
\input{sec/2_related}

\input{sec/3_method}

\input{sec/4_experiment}

\input{sec/5_conclusion}
{
    \small
    \bibliographystyle{ieeenat_fullname}
    \bibliography{main}
}


\end{document}

%% file: preamble.tex
\usepackage{algorithm}
\usepackage{algorithmic}
\usepackage{color, colortbl}
\usepackage{amsmath}
\usepackage{amssymb}
\usepackage{xcolor}
\usepackage{multirow}
\usepackage{graphicx}
\usepackage[accsupp]{axessibility} 

%% file: sec/0_abstract.tex
\begin{abstract}
Consistency learning is a central strategy to tackle unlabeled data in semi-supervised medical image segmentation (SSMIS), which enforces the model to produce consistent predictions under the perturbation. However, most current approaches solely focus on utilizing a specific single perturbation, which can only cope with limited cases, while employing multiple perturbations simultaneously is hard to guarantee the quality of consistency learning. In this paper, we propose an Adaptive Bidirectional Displacement (ABD) approach to solve the above challenge. Specifically, we first design a bidirectional patch displacement based on reliable prediction confidence for unlabeled data to generate new samples, which can effectively suppress uncontrollable regions and still retain the influence of input perturbations. Meanwhile, to enforce the model to learn the potentially uncontrollable content, a bidirectional displacement operation with inverse confidence is proposed for the labeled images, which generates samples with more unreliable information to facilitate model learning. Extensive experiments show that ABD achieves new state-of-the-art performances for SSMIS, significantly improving different baselines. Source code is available at \href{https://github.com/chy-upc/ABD}{https://github.com/chy-upc/ABD}.
\end{abstract}

%% file: sec/1_intro.tex
\section{Introduction}
Medical image segmentation derives from computer tomography (CT) or magnetic resonance imaging (MRI), which is crucial for various clinical applications~\cite{wang2019abdominal,zhao2023rcps}. Obtaining a large medical dataset with precise annotation to train segmentation models is challenging, as reliable annotations can only be provided by experts, which constrains the development of medical image segmentation algorithms and poses substantial challenges for further research and implementation~\cite{tajbakhsh2020embracing,zhang2023uncertainty}. To mitigate the burden of manual annotation and address these challenges, semi-supervised medical image segmentation (SSMIS)~\cite{sedai2019uncertainty,wu2022cross,bai2023bidirectional,zhang2023multi} is emerging as a practical approach to encourage segmentation models to learn from readily available unlabelled data in conjunction with limited labeled examples.

\begin{figure}[t]
\centering
\includegraphics[width=1\columnwidth]{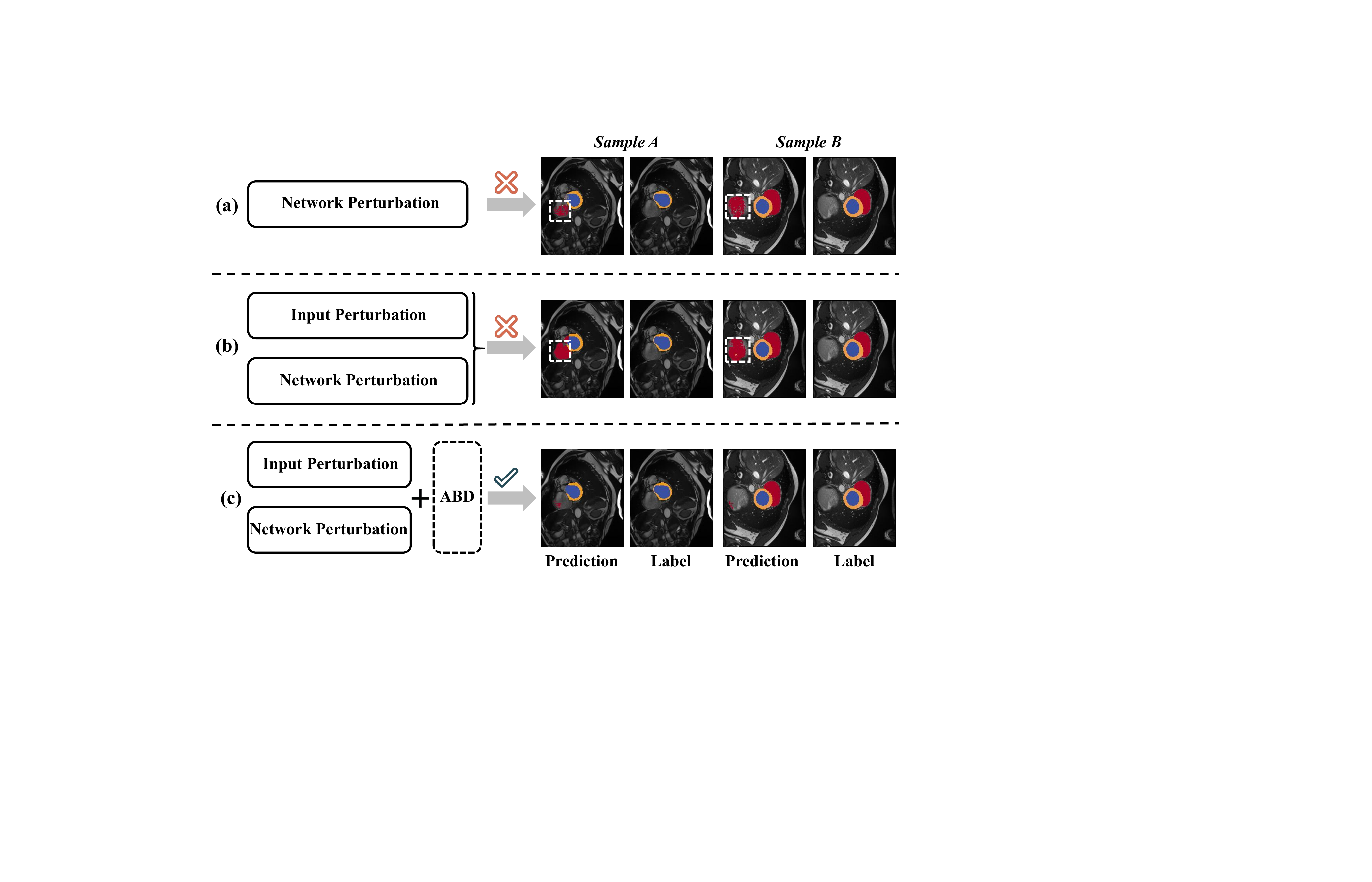} 
\caption{Illustration of prediction results. (a) Using only network perturbation; (b) Combining network perturbation with input perturbation; (c) Incorporating ABD on top of the two perturbations. Using a single perturbation has limitations while using multiple perturbations makes it uncontrollable. Introducing ABD greatly alleviates the issue and allows the model to perform significantly better. The white dashed boxes highlight the regions with wrong predictions.}
\label{fig1}
\end{figure}

Most recent approaches in SSMIS employ consistency learning~\cite{huang2022semi,zhang2023self,du2023coarse,lai2021semi} to make the decision boundary of the learned model located within the low-density boundary~\cite{jiao2022learning}. By ensuring consistent features or predictions under diverse perturbations, this strategy becomes one of the most effective solutions for learning from unlabelled data. According to the perturbation differences, consistency learning approaches can be divided into three categories: 1) Input perturbations, which mainly produce different inputs to the same model~\cite{berthelot2019mixmatch,shu2022cross,yang2023semi}, \eg, weak and strong data augmentation for the given image. 2) Feature perturbations~\cite{ouali2020semi,zheng2022double,li2021dual}, which mainly include feature noise, feature dropout, and context masking. 3) Network perturbations~\cite{ke2020guided,chen2021semi}, which focus on using different network architectures for the same input. To ensure the stability of consistency learning, most previous approaches \cite{tarvainen2017mean,zou2020pseudoseg,li2020transformation,xu2022learning} solely utilize one of the above perturbations, restricting the performance of the consistency learning and leading to imprecise decision boundary since the specific single perturbation can only handle limited cases, as shown in Fig.~\ref{fig1}(a).

Utilizing mixed or multiple perturbations presents a direct solution to solve the above problem. However, once added multiple perturbations, the consistency learning process is easily out-of-control, leading to restricted learning quality. For example, Cross Pseudo Supervision (CPS)~\cite{chen2021semi} is a widely-used technique in consistency learning that primarily applies network perturbation to produce two discriminate predictions for further consistency learning. If we directly add input perturbation by using weak and strong augmentation to the input training data, consistency learning will be ineffective. As shown in Fig.~\ref{fig1}(b), when the original input is replaced with weak and strong augmentation inputs, the CPS model incorrectly classifies the background as the foreground with consistency learning mechanism, indicating decreased performance when mixed perturbations are introduced.

To tackle the aforementioned challenges, we propose \textbf{Adaptive Bidirectional Displacement} (ABD) for SSMIS, as shown in Fig.~\ref{fig1}(c). Specifically, for each unlabelled image, the input perturbation is firstly applied to produce a pair of input images, \eg, weak augmentation image and strong augmentation image. Then we generate two confidence matrices (confidence rank maps) based on the predictions from the above two perturbed images. The confidence matrix assesses the certainty of predicted pixels belonging to various categories, thereby reflecting the model's reliability to different perturbations. Then, for any augmented image, its region with the lowest confidence rank is displaced with the region from the other augmented image that has the most similar output distribution with highly confident scores. We refer to this as an adaptive bidirectional displacement with reliable confidence (ABD-R). In this way, the newly generated image can remove the uncontrolled region and obtain complementary and approximate semantic information from another augmented image, which ensures consistent predictions of the model across different perturbations. Meanwhile, to enforce the model to learn those potentially uncontrollable regions, we incorporate inverse confidence for labeled data as an additional adaptive bidirectional displacement (ABD-I). For any labeled augmented image, the image regions with the highest confidence scores are displaced with the regions from another augmented image having the lowest confidence scores. This operation will strengthen the model to tackle uncontrollable regions. Combining these two strategies, our approach performs a novel input perturbation method, which can be directly applied to existing consistency learning approaches. Extensive experiments show that ABD achieves new state-of-the-art (SOTA) performances for SSMIS, significantly improving different baseline performances.

We summarize our main contributions as follows:

\begin{itemize}
\item 
We observed that the combination of different perturbations leads to instability in consistency learning. To address this issue, we propose Adaptive Bidirectional Displacement to enable the generation of semantically complementary data by replacing model inadaptable regions with credible regions, which assists the model in effectively correcting erroneous predictions and enhances the quality of consistency learning.

\item  
To take full advantage of labeled data, we propose an enhanced Adaptive Bidirectional Displacement that incorporates inverse confidence for labeled data to enforce the model to tackle those potentially uncontrollable regions.

\item 
The proposed method can be easily plug-and-play, which can be embedded into different approaches and enhance their performance. Extensive experiments are conducted to validate the feasibility of the method, resulting in significant improvements compared to previous SOTA approaches on different datasets.


\end{itemize}

%% file: sec/2_related.tex
\begin{figure*}[t]
\centering
\includegraphics[width=1.0\textwidth]{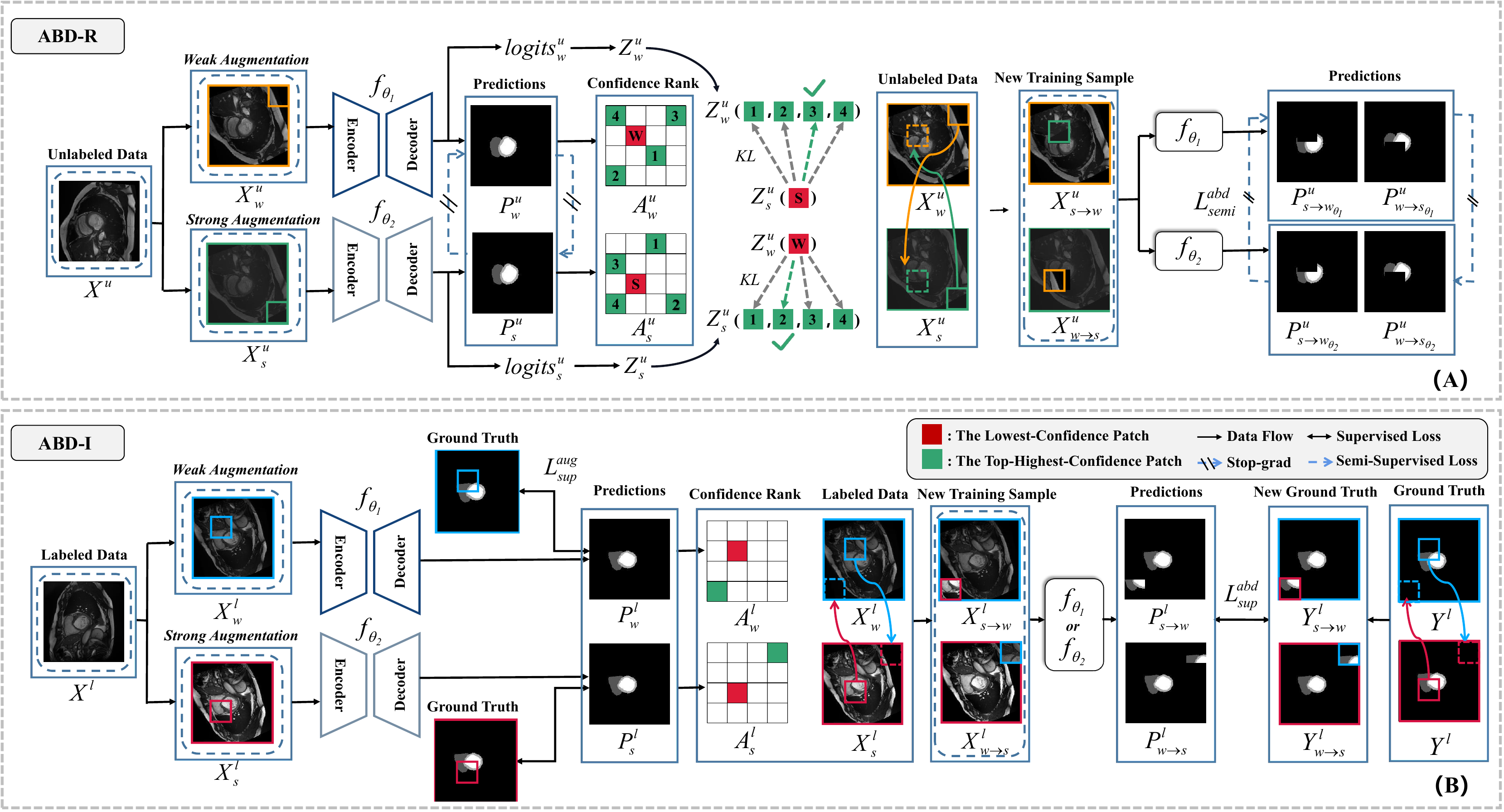} 
\caption{Overview of our adaptive bidirectional displacement framework. (A) For the unlabeled data, one image is subjected to weak and strong augmentations, resulting in two images that are separately input to two networks for cross-supervision. Then, based on the $A_w^u$, $Z_w^u$, $A_s^u$, and $Z_s^u$, the patches in the images are bidirectionally displaced, resulting in the formation of new samples ${X}_{s \rightarrow w}^u$ and ${X}_{w \rightarrow s}^u$. These new samples are further fed into the networks for cross-supervision. (B) For the labeled images, they are also subjected to both weak and strong augmentations, and their predictions are supervised by the labels. Afterward, based on the $A_w^l$ and $A_s^l$, inverse bidirectional patch displacement is performed on the images, resulting in the generation of new samples ${X}_{s \rightarrow w}^l$ and ${X}_{w \rightarrow s}^l$. Similarly, the labels undergo the same operation, leading to the creation of new labels $Y_{s\rightarrow w}^l$ and $Y_{w\rightarrow s}^l$. The new samples are then fed into the network, and their predictions are supervised by the new labels. Note that ABD-R and ABD-I are two parallel modules during training.} 
\label{fig2}
\end{figure*}

\section{Related Work}
\subsection{Consistency Learning in Semi-Supervised Medical Image Segmentation}
Consistency learning has been widely used in recent approaches for SSMIS. It aims to improve the performance of models by promoting consistent predictions for unlabeled data under different perturbations. According to the perturbation differences, consistency learning has three categories: input perturbation, feature perturbation, and network perturbation. Input perturbation is achieved through producing different inputs. For instance, Huang \emph{et al.}~\cite{huang2022semi} introduced cutout content loss and slice misalignment as perturbations in the input. In contrast, ST++~\cite{yang2022st++} involved strong augmentation on unlabeled images and directly utilizes the augmented samples for re-training. PseudoSeg~\cite{zou2020pseudoseg} was similar to FixMatch~\cite{sohn2020fixmatch} in leveraging pseudo-labels generated from weakly augmented images to supervise the predictions of strongly augmented images. BCP~\cite{bai2023bidirectional} encouraged the mixing of labeled and unlabeled images on the input level, enabling unlabeled data to learn comprehensive and general semantic information from labeled data in both directions. Meanwhile, feature perturbation and network perturbation are achieved by producing different features or outputs for the same input. Specifically, for feature perturbation, Mismatch~\cite{xu2022learning} introduced morphological feature perturbation, which is based on classic morphological operations. For network perturbation, CPS~\cite{chen2021semi} encouraged consistent predictions from different initialized networks with an input image. Mean Teacher~\cite{tarvainen2017mean} utilized the exponential moving average (EMA) technique to transfer semantic knowledge from the student network to the teacher network. However, existing methods only guarantee the consistency of predictions under a single perturbation, which greatly limits the scalability of consistency learning.

\subsection{Consistency Learning in Other Tasks}
Consistency learning~\cite{zhao2023augmentation,liang2023logic} is a major technique to tackle unlabeled data in semi-supervised learning. Several methods are proposed with consistency learning in various domains, including image segmentation~\cite{yang2023revisiting,ren2023viewco}, domain generalization~\cite{zhou2022uncertainty,zhou2023semi}, image classification~\cite{zheng2022simmatch,verma2022interpolation}, pre-train language model~\cite{he2022galaxy}, etc. For image segmentation, Revisiting UniMatch~\cite{yang2023revisiting} proposed a unified approach using dual-stream perturbations for semi-supervised semantic segmentation by combining input and feature perturbations. Additionally, ViewCo~\cite{ren2023viewco} introduced text-to-view consistency modeling, incorporating additional text to learn comprehensive segmentation masks. In domain generalization, Zhou \textit{et al.}~\cite{zhou2022uncertainty} exploited the latent uncertainty information of the unlabeled samples to design an uncertainty-guided consistency loss. StyleMatch~\cite{zhou2023semi} enforced prediction consistency between images from one domain and their style-transferred counterparts. In image classification, SimMatch~\cite{zheng2022simmatch} applied consistency regularization on both the semantic level and instance level to generate high-quality and reliable targets. ICT~\cite{verma2022interpolation} introduced the interpolation consistency training which encourages the prediction at an interpolation of unlabeled points to be consistent with the interpolation of the predictions at those points. In addition, in the pre-train language model, GALAXY~\cite{he2022galaxy} applied consistency regularization on all data to minimize the bi-directional KL divergence between model predictions.

%% file: sec/3_method.tex
\section{Method}
\subsection{Problem Setting}
In the semi-supervised segmentation, we utilize a labeled dataset $\mathcal{D}^l=\left\{\left(X_i^l, {Y}_i^l\right)\right\}_{i=1}^N$ along with an unlabeled dataset $\mathcal{D}^u=\left\{{X}_i^u\right\}_{i=N+1}^{M+N}$. Here, $X_i^l$ represents a labeled image and ${Y}_i^l$ represents its corresponding label. Similarly, ${X}_i^u$ denotes an unlabeled image. For convenience, \textbf{$i$ will be omitted in the following part.} It is worth emphasizing that the number of labeled images is significantly larger than the number of unlabeled images, \ie, $N \gg M$. 

\subsection{Overview}
The overall framework of our approach is shown in Fig.~\ref{fig2}, which can be divided into the following steps:

\begin{enumerate}
 \item For the input labeled image, we first apply input perturbation to generate two different samples, \eg, a weak augmentation image and a strong augmentation image, both of which are input to the model with network perturbation: the weak augmentation image is input to one network and the strong augmentation image is input to the other network to generate two corresponding predictions, using the provided label as supervision.
 
 \item For an unlabeled image, two predictions are obtained following the above pipeline. Then using our proposed ABD-R, the corresponding confidence matrices are used to perform bidirectional displacement between two augmented unlabeled images to generate new input samples.
 
 \item Meanwhile, to enforce learning from the potentially uncontrollable regions, the ABD-I strategy based on inverse confidence is applied to labeled data to generate two new samples with the corresponding labels.

 \item Finally, all generated new samples are input to the model to produce the corresponding predictions. The predictions of unlabeled data are used for cross-supervision, and the predictions of labeled data are supervised by the newly generated labels. 

\end{enumerate}

\subsection{Adaptive Bidirectional Displacement with Reliable Confidence}
Under various perturbations, the segmentation model produces unreliable predictions for unlabeled data. Imposing alignment using these predictions can render consistent learning ineffective. A direct solution is to remove the regions related to unreliable prediction. In practice, we first generate the confidence matrix based on the model's predictions. The confidence matrix measures the confidence of each predicted pixel belonging to different categories, it reflects the model’s reliability to different perturbations. Utilizing the confidence matrix, our ABD-R can generate a new training sample that exhibits more reliable regions.

Suppose ${X}_{w}^u\in \mathbb{R}^{3 \times H\times W}$ is an unlabeled medical image with weak augmentation. $X_s^u \in \mathbb{R}^{3 \times H\times W} $ is the same unlabeled medical image but with strong augmentation. After passing the model, we generate their corresponding outputs: 

\begin{equation}
\begin{aligned}
logits_{w}^u = f_{\theta_1}(X_{w}^u),\ logits_{s}^u = f_{\theta_2}(X_{s}^u),
\end{aligned}
\label{eq:logit}
\end{equation}
where $f_{\theta_1}$ and $f_{\theta_2}$ are two networks, which usually have different architecture or initialization to build the network perturbation. $logits_{w}^u$ and $logits_{s}^u$ are the logits outputs that correspond to $X_{w}^u$ and $X_{s}^u$, respectively. Applying softmax to these logits yields the corresponding prediction probability scores:


\begin{equation}
\resizebox{0.9\columnwidth}{!}{$
\begin{aligned}
P_{w}^u = \text{softmax}(logits_{w}^u), P_{s}^u = \text{softmax}(logits_{s}^u),
\end{aligned}
$}
\label{eq:P}
\end{equation}
where $P_{w}^u$ is the prediction of ${X}_{w}^u$ and $P_{s}^u$ is the prediction of $X_{s}^u$. After upsampling them to the same height and width with input, we can generate $P_{w}^u \in \mathbb{R}^{C \times H \times W}$ and $P_{s}^u \in \mathbb{R}^{C \times H \times W}$. $C$ is the number of classes. 

Then we divide ${X}_{w}^u$ into $K$ patches, each with a size of $k \times k$, denoting as ${X}_{w}^{u}=\{{X}_{w}^{u,j}\}_{j=1}^{K}$, where ${X}_{w}^{u,j}\in\mathbb{R}^{k\times k}$ and $K = \left( \frac{H}{k}\right) \times \left(\frac{W}{k} \right)$. For any patch ${X}_{w}^{u,j}$ ($j$ is the index of the patch), the corresponding logits outputs are represented as ${logits}_{w}^{u,j}\in\mathbb{R}^{C \times k\times k}$ and predictions are represented as ${P}_{w}^{u,j}\in\mathbb{R}^{C \times k\times k}$.

Then, for each patch of $X_w^u$, its logits scores is:
\begin{equation}
    Z_{w,c}^{u,j}= \frac{\sum\limits_{m=1}^{k \times k} logits_{w,c}^{u,j}(m)}{\left| k \times k\right|},
  \label{eq:Z_w^u}
\end{equation}
where $Z_{w,c}^{u,j}$ represents the average logits score of the $j$-th patch for the class $c$ and $c \in \{1,2, . . , C\}$, $logits_{w,c}^{u,j}(m)$ is the logit score of the class $c$ for the $m$-th pixel in the $j$-th patch. We regard $Z_{w}^{u,j}$ as the output distribution for the $j$-th patch of $X_w^u$.

Meanwhile, the corresponding confidence score for the $j$-th patch is computed as follows:
\begin{equation}
    A_w^{u,j}= \frac{\sum\limits_{m=1}^{k \times k} \max\limits_{c\in C}(P_{w,c}^{u,j}(m))}{\left| k \times k\right|},
  \label{eq:A_w^u}
\end{equation}
where $j$ is the index of the corresponding patch, $P_{w,c}^{u,j}(m)$ is the probability of the class $c$ for the $m$-th pixel in the patch. $\max(\cdot)$ is a maximum operator to select the highest confidence score for each pixel. $A_{w}^{u,j}$ is the average confidence score for the $j$-th patch, which measures the reliability of the corresponding patch.


Then the index of the patch with the lowest confidence for $X_w^u$ is computed as:

\begin{equation}
    \text{ind}_{w\text{-}min}^u= \underset{j \in K}{argmin} \left( A_{w}^{u,j}\right). \label{eq:ind_w_min}
\end{equation}

Simultaneously, using the confidence map $A_w^u$, the top $n$ highest confidence patches in $X_w^u$ are selected, and the corresponding index set is represented as:
\begin{equation}
    \text{Ind}_{w\text{-}top}^u = \left\{\text{ind}_{w\text{-}max^1}^u, \text{ind}_{w\text{-}max^2}^u, . . . , \text{ind}_{w\text{-}max^n}^u \right\},
\label{eq:ind_w_top}
\end{equation}
where $\text{Ind}_{w\text{-}top}^u$ is an index set that includes the indices of the selected top $n$ highest confidence patches. $\text{ind}_{w\text{-}max^{1\text{-}n}}^u$ is the selected indices, for example, $\text{ind}_{w\text{-}max^1}^u$ is the index of the patch that has the highest (top 1) confidence score.

Similarly, following the above operations, we divide ${X}_{s}^u$ into $K$ patches, the corresponding logits and confidence score are represented as ${Z}_{w}^{u,j}$ and ${A}_{s}^{u,j}$, respectively. After processing from Eq.~(\ref{eq:ind_w_min}) to Eq.~(\ref{eq:ind_w_top}), the indices of the top-highest and the lowest confidence for ${X}_{s}^u$ are generated, representing as $\text{Ind}_{s\text{-}top}^u$ and $\text{ind}_{s\text{-}min}^u$, respectively. And $\text{Ind}_{s\text{-}top}^u = \left\{\text{ind}_{s\text{-}max^1}^u, \text{ind}_{s\text{-}max^2}^u, . . . , \text{ind}_{s\text{-}max^n}^u \right\}.$

To achieve controlled effects of the mixed perturbations and enhance semantic coherence of the displacement operation, we select the most similar patch from the obtained index sets ($\text{Ind}_{w\text{-}top}^u$ and $\text{Ind}_{s\text{-}top}^u$) by calculating the difference of the output distribution between the selected top-$n$ highest confidence patches in one image and the lowest confidence patch in the other image:
\begin{equation}
\resizebox{0.9\columnwidth}{!}{$
\begin{aligned}
    \text{ind}_{w\text{-}max^{s}} = \underset{i \in n}{argmin} \left( kl_{div}(Z_{w}^{u,\text{ind}_{w\text{-}max^i}^u}, Z_{s}^{u,\text{ind}_{s\text{-}min}^u})\right) \\
    \text{ind}_{s\text{-}max^{s}} = \underset{i \in n}{argmin} \left( kl_{div}(Z_{s}^{u,\text{ind}_{s\text{-}max^i}^u}, Z_{w}^{u,\text{ind}_{w\text{-}min}^u})\right) \\
\end{aligned},\label{eq:kl_div}
$}
\end{equation}
where $kl_{div}$ is to compute the KL divergence~\cite{kullback1951information}, \eg, $kl_{div}(Z_{w}^*, Z_{s}^*)) = \text{softmax}\left(Z_{w}^* \right) \log \frac{\text{softmax}\left(Z_{w}^*\right)}{\text{softmax}\left(Z_{s}^*\right)}$, which reflects the difference in the output distribution.
$\text{ind}_{w\text{-}max^{s}}$ and $\text{ind}_{s\text{-}max^{s}}$ represent the indices of final selected patches.

Finally, the bidirectional displacement operation is performed for $X_s^u$ and $X_w^u$. The patch of one augmented image with the lowest confidence score is displaced with the final selected patch from the other augmented image:

\begin{equation}
      X_{s \rightarrow w}^{u,j} =
      \begin{cases} 
        X_{s}^{u,\text{ind}_{s\text{-}max^s}^u}, &\text{ if } j=\text{ind}_{w\text{-}min}^u \\ 
        X_{w}^{u,j}, &\text{else}
      \end{cases},
\end{equation}

\begin{equation}
      X_{w \rightarrow s}^{u,j} =
      \begin{cases} 
        X_{w}^{u,\text{ind}_{w\text{-}max^s}^u}, &\text{ if } j=\text{ind}_{s\text{-}min}^u \\ 
        X_{s}^{u,j}, &\text{else}
      \end{cases}.
\end{equation}

After removing the patching operation and reshaping them to the image, two new samples $X_{s \rightarrow w}^{u}$ and $X_{w \rightarrow s}^{u}$ are generated, which are then input to two networks $f_{\theta_1}$ and $f_{\theta_2}$ for cross-supervision, as shown in Eq.~(\ref{eq:l_semi_abd}).

This approach allows the two networks to adapt to each other's input perturbations, resulting in higher quality and more reliable pseudo-labels. As a cross-supervised process for the unlabeled data, it also reduces the likelihood of one model making erroneous predictions, thereby preventing the degradation of the other model's training process.

\subsection{Adaptive Bidirectional Displacement with Inverse Confidence}

In the above ABD-R module, unlabeled images are manipulated to create new samples by replacing the potentially uncontrollable patches with new patches, it does not directly learn from the original regions. To enforce the model to learn from these potentially uncontrollable regions, we propose Adaptive Bidirectional Displacement with Inverse Confidence (ABD-I) for the labeled images to strengthen the learning for uncontrollable regions.

Specifically, for a labeled image $X^{l}$, we perform a strong augmentation to obtain a sample $X_{s}^l$ and a weak augmentation to obtain a sample $X_{w}^l$. Let $X_{w}^l$ as an example, we execute Eq.~(\ref{eq:logit}) to Eq.~(\ref{eq:ind_w_top}) on $X_{w}^l$ to get its region index, \ie, $\text{ind}_{w\text{-}max^1}^l$ and $\text{ind}_{w\text{-}min}^l$. Similarly, we repeat the above operation for the strong augmented image $X_{s}^l$ to get $\text{ind}_{s\text{-}max^1}^l$ and $\text{ind}_{s\text{-}min}^l$. 
Note that $\text{ind}_{w\text{-}max^1}^l$ and $\text{ind}_{s\text{-}max^1}^l$ are the indices of the patches that have the highest (top 1) confidence score, we directly use them rather than selecting from candidates since these regions are the most potentially controllable region and we do not need to consider the semantic coherence with the given annotations.  $\text{ind}_{w\text{-}min}^l$ and $\text{ind}_{s\text{-}min}^l$ share the same definition with Eq.~(\ref{eq:ind_w_min}).

Then ABD-I is used, the region of one image with the highest confidence scores is displaced with the region from the other image having the lowest confidence scores:

\begin{equation}
      X_{s \rightarrow w}^{l,j} =
      \begin{cases} 
        X_{s}^{l,\text{ind}_{s\text{-}min}^l}, &\text{ if } j=\text{ind}_{w\text{-}max^1}^l \\ 
        X_{w}^{l,j}, &\text{else}
      \end{cases}\label{eq:X_sw^lj},
\end{equation}

\begin{equation}
      X_{w \rightarrow s}^{l,j} =
      \begin{cases} 
        X_{w}^{l,\text{ind}_{w\text{-}min}^l}, &\text{ if } j=\text{ind}_{s\text{-}max^1}^l \\ 
        X_{s}^{l,j}, &\text{else}
      \end{cases}.\label{eq:X_ws^lj}
\end{equation}

Using Eq.~(\ref{eq:X_sw^lj}) and Eq.~(\ref{eq:X_ws^lj}), the new samples $X_{s \rightarrow w}^{l}$ and $X_{w \rightarrow s}^{l}$ are generated. To effectively supervise the displaced samples, the same displacement is applied to the original label. Suppose $Y^l$ is the label of images $X^l_s$ and $X^l_w$, we also divide $Y^l$ into $K$ patches, denoting as $Y^{l,j}=\{{Y}^{l,j}\}_{j=1}^{K}$. The corresponding displacement operation is expressed as:
\begin{equation}
      Y_{s \rightarrow w}^{l,j} =
      \begin{cases} 
        Y^{l,\text{ind}_{s\text{-}min}^l}, &\text{ if } j=\text{ind}_{w\text{-}max^1}^l \\ 
        Y^{l,j}, &\text{else}
      \end{cases},
\end{equation}

\begin{equation}
      Y_{w \rightarrow s}^{l,j} =
      \begin{cases} 
        Y^{l,\text{ind}_{w\text{-}min}^l}, &\text{ if } j=\text{ind}_{s\text{-}max^1}^l \\ 
        Y^{l,j}, &\text{else}
      \end{cases}.
\end{equation}

Finally $Y_{s \rightarrow w}^{l}$ and $Y_{w \rightarrow s}^{l}$ are used to supervise the generated samples $X_{s \rightarrow w}^{l}$ and $X_{w \rightarrow s}^{l}$, as shown in Eq.~(\ref{eq:l_sup_abd}).

\subsection{Loss Functions}
The overall loss function consists of two parts, the supervised loss $\mathcal{L}_{sup}$ for labeled data and the unsupervised loss $\mathcal{L}_{semi}$ for unlabeled data, defined as:
\begin{equation}
\begin{aligned}
  \mathcal{L}&=\mathcal{L}_{sup} + \lambda \mathcal{L}_{semi} \\
  &=\left( \mathcal{L}_{sup}^{aug} + \mathcal{L}_{sup}^{abd} \right) + \lambda \left( \mathcal{L}_{semi}^{aug} + \mathcal{L}_{semi}^{abd} \right)
  \label{eq:l_sup}
\end{aligned}
\end{equation}
where $\lambda$ is a weight that balances different losses, which gradually increase with iterations increases based on the Gaussian warming up: $\lambda= 0.1 \cdot e^{-5 \times \left(1-{t} / {t_{\text {total}}}\right)^2}$, $t$ is the current iteration and $t_{\text {total}}$ is the total iteration.

The $\mathcal{L}_{sup}$ consists of $ \mathcal{L}_{sup}^{aug}$ and $\mathcal{L}_{sup}^{abd}$. $\mathcal{L}_{sup}^{aug}$ is used for the original augmented labeled images, $\mathcal{L}_{sup}^{abd}$ is used for the newly generated labeled samples. 

The loss $\mathcal{L}_{sup}^{aug}$ is computed as:
\begin{equation}
\resizebox{0.85\columnwidth}{!}{$
\begin{aligned}
    \mathcal{L}_{sup}^{aug} = \frac{1}{2} \times (\mathcal{L}_{ce}(f_{\theta_1}(X_{w}^l),Y^l) + \mathcal{L}_{dice}(f_{\theta_1}(X_{w}^l),Y^l))\\
  + \frac{1}{2} \times (\mathcal{L}_{ce}(f_{\theta_2}(X_{s}^l),Y^l) + \mathcal{L}_{dice}(f_{\theta_2}(X_{s}^l),Y^l))
  \label{eq:1}
\end{aligned}
$}
\end{equation}
where $\mathcal{L}_{ce}$, $\mathcal{L}_{dice}$ are the cross-entropy loss and dice loss \cite{milletari2016v}, respectively. $Y^l$ is the label.

The loss $\mathcal{L}_{sup}^{abd}$ is computed as:
\begin{equation}
\begin{aligned}
    \mathcal{L}_{sup}^{abd} = \frac{1}{2} \times (\mathcal{L}_{ce}(f_{\theta_1}(X_{s \rightarrow w}^l),Y_{s \rightarrow w}^l) \\
    +\mathcal{L}_{dice}(f_{\theta_1}(X_{s \rightarrow w}^l),Y_{s \rightarrow w}^l))  \\
    +\frac{1}{2} \times (\mathcal{L}_{ce}(f_{\theta_2}(X_{w \rightarrow s}^l),Y_{w \rightarrow s}^l)\\
    +\mathcal{L}_{dice}(f_{\theta_2}(X_{w \rightarrow s}^l),Y_{w \rightarrow s}^l))
  \label{eq:l_sup_abd}
\end{aligned}
\end{equation}

The unsupervised loss $\mathcal{L}_{semi}$ consists of $\mathcal{L}_{semi}^{aug}$ and $\mathcal{L}_{semi}^{abd}$. $\mathcal{L}_{semi}^{aug}$ is used for the original augmented unlabeled images, $\mathcal{L}_{semi}^{abd}$ is for the newly generated unlabeled images.

The loss function $\mathcal{L}_{semi}^{aug}$ is computed as follows:
\begin{equation}
\begin{aligned}
  \mathcal{L}_{semi}^{aug} = \mathcal{L}_{dice}(f_{\theta_1}(X_{w}^u),\underset{c \in C}{argmax}(P^u_{s,c}))\\
  +\mathcal{L}_{dice}(f_{\theta_2}(X_{s}^u),\underset{c \in C}{argmax}(P^u_{w,c}))
  \label{eq:semi_aug}
  \end{aligned}
\end{equation}
where $P_{s,c}^u$ and $P_{w,c}^u$ are the predictions in Eq~(\ref{eq:P}). 

\begin{table*}[h]
\caption{Comparisons with other methods on the ACDC test set, ``\textbf{Ours}-ABD (Cross Teaching)" and ``\textbf{Ours}-ABD (BCP)" represent the baseline is Cross Teaching ~\cite{luo2022semi} and BCP ~\cite{bai2023bidirectional} respectively.}\label{tab:ACDC}
\centering
\resizebox{\textwidth}{!}{$
\begin{tabular}{c|cc|cccc}
\hline
\multirow{2}{*}{Method} & \multicolumn{2}{c|}{Scans used} & \multicolumn{4}{c}{Metrics} \\
\cline{2-7}
& Labeled & Unlabeled & DSC$\uparrow$ & Jaccard$\uparrow$ & 95HD$\downarrow$ & ASD$\downarrow$ \\
\hline
\multirow{3}{*}{U-Net (MICCAI'2015) ~\cite{ronneberger2015u}} & 3(5\%) & 0 & 47.83 & 37.01 & 31.16 & 12.62 \\
& 7(10\%) & 0 & 79.41 & 68.11 & 9.35 & 2.70 \\
& 70(All) & 0 & 91.44 & 84.59 & 4.30 & 0.99 \\
\hline
DTC (AAAI'2021) ~\cite{luo2021semi} & \multirow{9}{*}{3(5\%)} & \multirow{9}{*}{67(95\%)} & 56.90 & 45.67 & 23.36 & 7.39 \\
URPC (MICCAI'2021) ~\cite{luo2021efficient} & & & 55.87 & 44.64 & 13.60 & 3.74 \\
MC-Net (MICCAI'2021) ~\cite{wu2021semi} & & & 62.85 & 52.29 & 7.62 & 2.33 \\
SS-Net (MICCAI'2022) ~\cite{wu2022exploring} & & & 65.83 & 55.38 & 6.67 & 2.28 \\
SCP-Net (MICCAI'2023) ~\cite{zhang2023self} & & & 87.27 & - & - & 2.65 \\
\cline{1-1} \cline{4-7} 
Cross Teaching (Reported) (MIDL'2022) ~\cite{luo2022semi} & & & 65.60 & - & 16.2 & - \\
BCP (CVPR'2023) \cite{bai2023bidirectional} & & & 87.59 & 78.67 & 1.90 & 0.67 \\
\cline{1-1} \cline{4-7}
\textbf{Ours}-ABD (Cross Teaching) & & & 86.35 & 76.73 & 4.12 & 1.22 \\
\textbf{Ours}-ABD (BCP) & & & \textbf{88.96} & \textbf{80.70} & \textbf{1.57} & \textbf{0.52} \\
\hline
DTC (AAAI'2021) ~\cite{luo2021semi} & \multirow{11}{*}{7(10\%)} & \multirow{11}{*}{63(90\%)} & 84.29 & 73.92 & 12.81 & 4.01 \\
URPC (MICCAI'2021) ~\cite{luo2021efficient} & & & 83.10 & 72.41 & 4.84 & 1.53 \\
MC-Net (MICCAI'2021) ~\cite{wu2021semi} & & & 86.44 & 77.04 & 5.50 & 1.84 \\
SS-Net (MICCAI'2022) ~\cite{wu2022exploring} & & & 86.78 & 77.67 & 6.07 & 1.40 \\ 
SCP-Net (MICCAI'2023) ~\cite{zhang2023self} & & & 89.69 & - & - & 0.73 \\
PLGCL (CVPR'2023) ~\cite{basak2023pseudo} & & & 89.1 & - & 4.98 & 1.80 \\
\cline{1-1} \cline{4-7} 
Cross Teaching (Reported) (MIDL'2022) ~\cite{luo2022semi} & & & 86.40 & - & 8.60 & - \\
Cross Teaching (Reproduced) & & & 86.45 & 77.02 & 6.30 & 1.86 \\
BCP (CVPR'2023) ~\cite{bai2023bidirectional} & & & 88.84 & 80.62 & 3.98 & 1.17 \\
\cline{1-1} \cline{4-7}
\textbf{Ours}-ABD (Cross Teaching) & & & 88.52 & 79.97 & 5.06 & 1.43  \\
\textbf{Ours}-ABD (BCP) & & & \textbf{89.81} & \textbf{81.95} & \textbf{1.46} & \textbf{0.49} \\
\hline
\end{tabular}%
$}
\end{table*}

The loss function $\mathcal{L}_{semi}^{abd}$ is defined as:
\begin{equation}
\begin{aligned}
  \mathcal{L}_{semi}^{abd} = \mathcal{L}_{dice}(f_{\theta_1}(X_{s\rightarrow w}^u),\underset{c \in C}{argmax}(P_{s\rightarrow w_{\theta_2},c}^u)) \\
  + \mathcal{L}_{dice}(f_{\theta_2}(X_{s\rightarrow w}^u),\underset{c \in C}{argmax}(P_{s\rightarrow w_{\theta_1},c}^u)) \\
  + \mathcal{L}_{dice}(f_{\theta_1}(X_{w\rightarrow s}^u),\underset{c \in C}{argmax}(P_{w\rightarrow s_{\theta_2},c}^u)) \\
  + \mathcal{L}_{dice}(f_{\theta_2}(X_{w\rightarrow s}^u),\underset{c \in C}{argmax}(P_{w\rightarrow s_{\theta_1},c}^u)) 
  \label{eq:l_semi_abd}
\end{aligned}
\end{equation}
where $P_{s\rightarrow w_{\theta_1}}^u = f_{\theta_1}(X_{s\rightarrow w}^u)$ and $P_{s\rightarrow w_{\theta_2}}^u = f_{\theta_2}(X_{s\rightarrow w}^u)$, which are the predictions from $f_{\theta_1}$ and $f_{\theta_2}$, respectively. Similarly, $P_{w\rightarrow s_{\theta_1}}^u$ and $P_{w\rightarrow s_{\theta_2}}^u$ are obtained using the same operation.

%% file: sec/4_experiment.tex
\section{Experiments}
\subsection{Dataset and Evaluation Metrics}

\textbf{ACDC dataset:} The ACDC dataset~\cite{bernard2018deep} comprises 200 annotated short-axis cardiac cine-MR images from a cohort of 100 patients with four classes. 2D segmentation is more common than direct 3D segmentation~\cite{bai2017semi}. The data split remains 70 patients' scans for training, 10 for validation, and 20 for testing. Following previous approaches, four evaluation metrics are chosen: \textit{Dice Similarity Coefficient} (DSC), \textit{Jaccard}, \textit{95\% Hausdorff Distance} (95HD) and \textit{Average Surface Distance} (ASD).

\textbf{PROMISE12 dataset:} The PROMISE12 dataset ~\cite{litjens2014evaluation} was made available for the MICCAI 2012 prostate segmentation challenge. MRI of 50 patients with various diseases was acquired at different locations. All 3D scans are converted into 2D slices. Following previous approaches, DSC and ASD are used as evaluation metrics.

\subsection{Implementation Details} 

We evaluate our approach on two baselines: Cross Teaching~\cite{luo2022semi} and BCP~\cite{bai2023bidirectional}.

Cross Teaching~\cite{luo2022semi} is a cross-supervision framework, which utilizes two networks to perform the network perturbation: $f_{\theta_1}$ and $f_{\theta_2}$, representing UNet ~\cite{ronneberger2015u} and Swin-UNet~\cite{cao2022swin}, respectively.
We add the input perturbation using weak and strong data augmentation to provide two kinds of inputs. For weak data augmentation, random rotation and flipping are used. For strong data augmentations, colorjitter, blur~\cite{chen2020improved} and Cutout~\cite{yang2022st++} are used. During training, all inputs are cropped to $224\times224$ and divided into $K=16$ patches. The networks are trained with a batch size of 16, including 8 labeled slices and 8 unlabeled slices. 

BCP ~\cite{bai2023bidirectional} is a mean-teacher framework, which applies weak and strong augmentation to perform input perturbation. We add the network perturbation by providing two student networks with different initializations. Our ABD is used on these two student networks following the above design. During training, all inputs are cropped to $256\times256$ and divided into $K=16$ patches. 

All other settings follow the default settings with original Cross Teaching~\cite{luo2022semi} and BCP~\cite{bai2023bidirectional}.

\begin{table}[htb]
\caption{Comparisons with state-of-the-art semi-supervised segmentation methods on the PROMISE12 test set.}\label{tab:PROMISE12}
\centering
\resizebox{\columnwidth}{!}{$
\begin{tabular}
{c|cc|cc}
\hline
\multirow{2}{*}{Method} & \multicolumn{2}{c|}{Scans used} & \multicolumn{2}{c}{Metrics} \\
\cline{2-5}
& Labeled & Unlabeled & DSC$\uparrow$ & ASD$\downarrow$ \\
\hline
\multirow{2}{*}{U-Net~\cite{ronneberger2015u}} & 7(20\%) & 0 & 60.88 & 13.87 \\
& 35(100\%) & 0 & 84.76 & 1.58 \\
\hline
CCT~\cite{ouali2020semi} & \multirow{6}{*}{7(20\%)} & \multirow{6}{*}{28(80\%)} & 71.43 & 16.61 \\
URPC~\cite{luo2021efficient} & & & 63.23 & 4.33 \\
SS-Net~\cite{wu2022exploring} & & & 62.31 & 4.36 \\
SLC-Net~\cite{liu2022semi} & & & 68.31 & 4.69 \\
SCP-Net~\cite{zhang2023self} & & & 77.06 & 3.52 \\ \cline{1-1} \cline{4-5}
\textbf{Ours}-ABD & & & \textbf{82.06} & \textbf{1.33} \\ 
\hline
\textbf{Ours}-ABD & 3(10\%) & 32(90\%) & \textbf{81.81} & \textbf{1.46} \\
\hline
\end{tabular}
$}
\end{table}

\subsection{Comparison with Sate-of-the-Art Methods}
\textbf{ACDA dataset:} Table~\ref{tab:ACDC} shows the average performance on the ACDC dataset using 5\% and 10\% labeled ratios. For the 5\% label ratio, our ABD achieves state-of-the-art performance compared to recent methods, surpassing the baseline Cross Teaching by a large margin. BCP is one of the newly released methods, it achieved high performance on the ACDC dataset. By incorporating our method into BCP, \ie, \textbf{Ours}-ABD (BCP), it has a noticeable improvement, which highlights the flexibility and scalability of our approach. Specifically, we observe a 20.75\% DSC improvement compared to Cross Teaching and a 1.37\% DSC improvement compared to BCP. Considering the 10\% labeled ratio, our ABD outperforms SCP-Net~\cite{zhang2023self} in all evaluation metrics and achieves a new state-of-the-art performance. It demonstrates that ABD mitigates the uncontrolled effects caused by the mixture of multiple perturbations and validates the mixed perturbations could expand the upper bound of consistency learning.

\begin{figure}[t]
\centering
\includegraphics[width=\columnwidth]{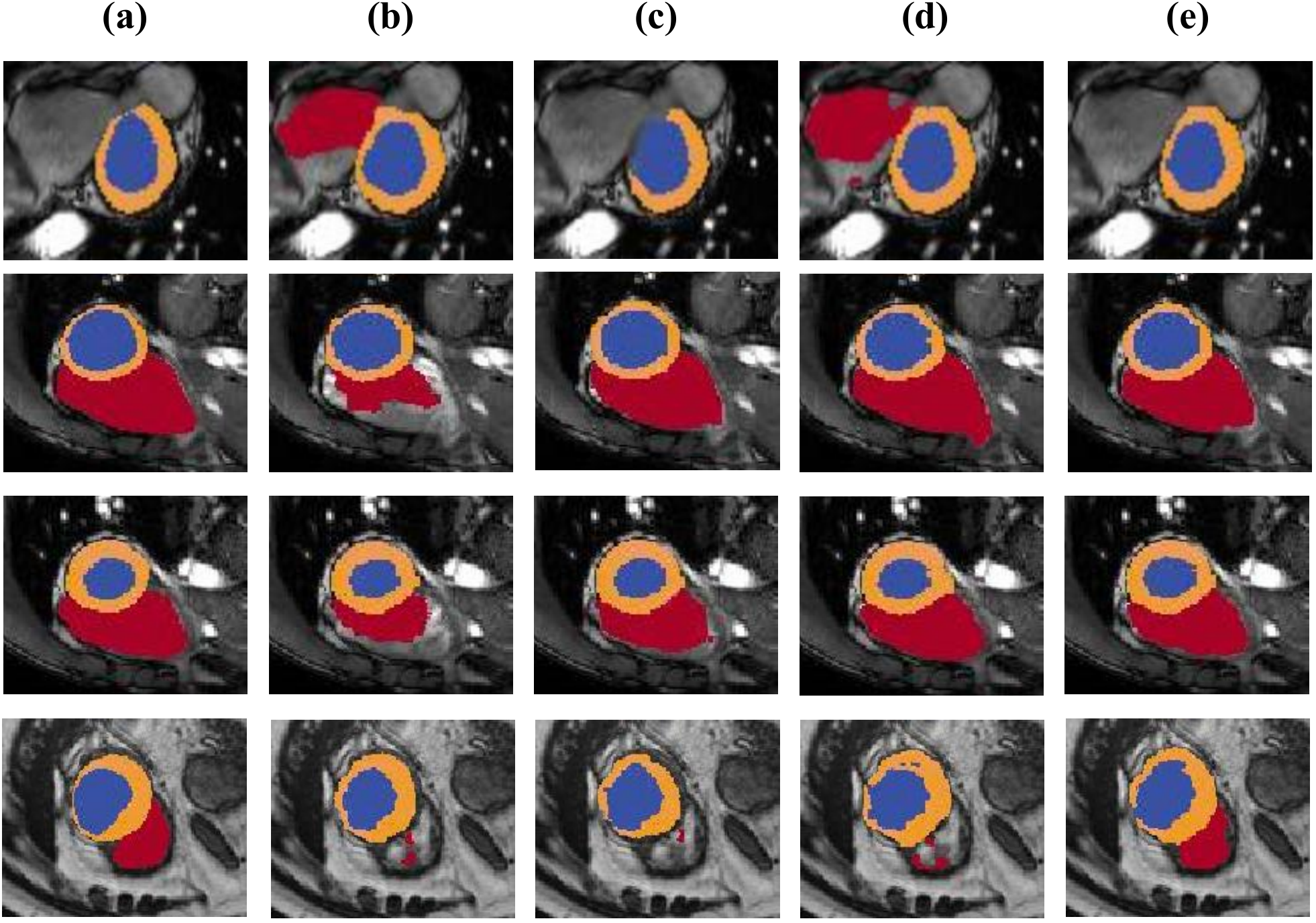} 
\caption{Visualization of segmentation results on ACDC dataset with 10\% labeled data. (a) Ground-truth. (b) Cross Teaching results. (c) \textbf{Ours}-ABD (Cross Teaching) results. (d) BCP results. (e) \textbf{Ours}-ABD (BCP) results. Best viewed in color on the screen.}
\label{fig3}
\end{figure}

Fig.~\ref{fig3} gives some qualitative results, our method effectively suppresses areas that are inaccurately segmented by Cross Teaching and BCP, while accurately segmenting the foregrounds that are ignored by the two methods. 

\textbf{PROMISE12 dataset:} Following SS-Net~\cite{wu2022exploring}, we conduct experiments with 20\% labeled data. We compare ABD with CCT~\cite{ouali2020semi}, URPC~\cite{luo2021efficient}, SS-Net~\cite{wu2022exploring}, SLC-Net~\cite{liu2022semi} and SCP-Net~\cite{zhang2023self}. As shown in Table~\ref{tab:PROMISE12}, Ours-ABD surpasses all other approaches. Compared to the recently proposed SCP-Net, ABD brings a 5.0\% DSC increase. 

\subsection{Ablation Studies}
We select Cross Teaching~\cite{luo2022semi} as the baseline. All experiments are conducted on the ACDC dataset with 10\% labeled data unless otherwise stated.

\textbf{Effectiveness of each module in ABD:} Table~\ref{tab:module} demonstrates the effectiveness of each module in ABD by progressively adding ABD-R and ABD-I. It can be observed that the introduction of input perturbation (IP) decreases the performance of the baseline, indicating that the mixed perturbation easily leads to out-of-control. Incorporating ABD-R and ABD-I brings increased performance. By incorporating both ABD-R and ABD-I into the baseline, the model has the best result, reaching 88.52\% DSC and surpassing the baseline in 2.07\%. The improvement indicates that the two modules are complementary which is consistent with the design targets. ABD-R enhances the upper limit of consistency learning for mixed perturbations, and ABD-I enables the model to learn potentially uncontrollable regions, thereby the combination can learn more semantics.

\begin{table}[!t]\caption{Effectiveness of ABD-R and ABD-I modules. ``base" means baseline is the Cross Teaching. ``IP" means adding perturbation to the input.}\label{tab:module}
\centering
\resizebox{\columnwidth}{!}{$
\begin{tabular}
{cccc|cccc}
\hline
Base & IP & ABD-R & ABD-I & DSC$\uparrow$ & Jaccard$\uparrow$ & 95HD $\downarrow$ & ASD$\downarrow$ \\
\hline
\checkmark & & & & 86.45 & 77.02 & 6.30 & 1.86\\
\checkmark & \checkmark & & & 86.25 & 76.69 & 5.44 & 1.72\\

\checkmark & \checkmark & \checkmark & & 87.42 & 78.37 & 5.23 & 1.68 \\
\checkmark & \checkmark& &\checkmark & 87.20 & 78.07 & 6.06 & 1.96\\
\checkmark & \checkmark &\checkmark &\checkmark & \textbf{88.52} & \textbf{79.97} & \textbf{5.06} & \textbf{1.43} \\
\hline
\end{tabular}%
$}
\end{table}

\textbf{Displacement Strategies:} Table~\ref{tab:strategies} illustrates the impact of different displacement strategies. The default displacement strategy (denoted as ``Reliable") in our ABD-R module is displacing the region with the top-highest confidence from one image with the region of lowest confidence from the other image. ``Random" refers to randomly selecting patches for displacement. ``Same" means the displacement strategy is performed by displacing the region that has the highest confidence from one image with the corresponding region from the other image. ``Same+Reliable" represents that in each iteration, we randomly select a strategy from the ``Same" or ``Reliable". The ``Same" generates samples that have more uncontrolled regions and yields unsatisfactory results. While the ``Same+Reliable" alleviates the unfavorable results, it is still not as effective as our approach.

\begin{table}[!t]
\caption{Influence of the displacement strategies in ABD-R.}\label{tab:strategies}
\centering
\begin{tabular}
{c|cccc}
\hline
Strategy & DSC$\uparrow$ & Jaccard$\uparrow$ & 95HD$\downarrow$ & ASD$\downarrow$ \\
\hline
Random & 86.55 & 77.07 & 6.13 & 1.74 \\
Same & 87.22 & 78.04 & 5.61 & 1.50 \\
Same+Reliable & 87.38 & 78.06 & \textbf{4.56} & 1.69 \\
\textbf{Reliable} & \textbf{87.42} & \textbf{78.37} & 5.23 & \textbf{1.68} \\
\hline
\end{tabular}
\end{table}

\textbf{Selection of Patch Number:} Table~\ref{tab:size} indicates the impact of the patch number. $K$ represents the number of patches. An optimal result is achieved when $K=16$. 


\begin{table}[!t]
\caption{Ablation study of patch number $K$.} \label{tab:size}
\centering
\begin{tabular}
{c|cccc}
\hline
$K$ & DSC$\uparrow$ & Jaccard$\uparrow$ & 95HD $\downarrow$ & ASD $\downarrow$ \\
\hline
4  & 87.39 & 78.35 & \textbf{5.06} & 1.69  \\
16 & \textbf{88.52} & \textbf{79.97} & \textbf{5.06} & \textbf{1.43} \\
64 & 87.54 & 78.57 & 5.52 & 1.88 \\ 
\hline
\end{tabular}

\end{table}

\textbf{Choice of Strong Augmentation:} Table~\ref{tab:aug} illustrates the influence of three different augmentation methods: colorjitter, blur, and cutout. Through comparison, combining colorjitter and cutout produces better performance.

\begin{table}[!t]
\caption{Effectiveness of the strong data augmentation.}\label{tab:aug}
\centering
\resizebox{\columnwidth}{!}{$
\begin{tabular}{ccc|cccc}
\hline
Cutout & Colorjitter & Blur & DSC$\uparrow$ & Jaccard$\uparrow$ & 95HD$\downarrow$ & ASD$\downarrow$ \\
\hline
\checkmark & & & 88.23 & 79.53 & 5.90 & \textbf{1.40} \\
& \checkmark & & 88.03 & 79.34 & 7.15 & 1.76 \\
& & \checkmark & 87.76 & 78.76 & 7.28 & 1.61 \\
\checkmark & \checkmark & & \textbf{88.52} & \textbf{79.97} & \textbf{5.06} & 1.43 \\
\checkmark &\checkmark &\checkmark & 87.83 & 79.02 & 6.14 & 1.87 \\
\hline
\end{tabular}
$}
\end{table}

%% file: sec/5_conclusion.tex
\section{Conclusion}
In this paper, we proposed an adaptive bidirectional displacement (ABD) for semi-supervised medical image segmentation. Our key idea is to mitigate the constraints of mixed perturbations on consistency learning, thereby enhancing the upper limit of consistency learning. To achieve this, we designed two novel modules in our ABD: an ABD-R module reduces the uncontrolled regions in unlabeled samples and captures comprehensive semantic information from input perturbations, and an ABD-I module enhances the learning capacity to uncontrollable regions in labeled samples to compensate for the deficiencies of ABD-R. With the cooperation of two modules, our method achieves state-of-the-art performance and is easily embedded into different methods. In the future, we will design a patch adaptive displacement strategy to tackle more complicated cases. 

\textbf{Acknowledge:} This work was supported by National Natural Science Foundation of China (No. 62301613), the Taishan Scholar Program of Shandong (No. tsqn202306130), the Shandong Natural Science Foundation (No. ZR2023QF046), Qingdao Postdoctoral Applied Research Project (No. QDBSH20230102091) and Independent Innovation Research Project of China University of Petroleum (East China) (No. 22CX06060A).

%% file: main.bbl
\begin{thebibliography}{51}
\providecommand{\natexlab}[1]{#1}
\providecommand{\url}[1]{\texttt{#1}}
\expandafter\ifx\csname urlstyle\endcsname\relax
  \providecommand{\doi}[1]{doi: #1}\else
  \providecommand{\doi}{doi: \begingroup \urlstyle{rm}\Url}\fi

\bibitem[Bai et~al.(2017)Bai, Oktay, Sinclair, Suzuki, Rajchl, Tarroni,
  Glocker, King, Matthews, and Rueckert]{bai2017semi}
Wenjia Bai, Ozan Oktay, Matthew Sinclair, Hideaki Suzuki, Martin Rajchl,
  Giacomo Tarroni, Ben Glocker, Andrew King, Paul~M Matthews, and Daniel
  Rueckert.
\newblock Semi-supervised learning for network-based cardiac mr image
  segmentation.
\newblock In \emph{Medical Image Computing and Computer-Assisted Intervention},
  pages 253--260. Springer, 2017.

\bibitem[Bai et~al.(2023)Bai, Chen, Li, Shen, and Wang]{bai2023bidirectional}
Yunhao Bai, Duowen Chen, Qingli Li, Wei Shen, and Yan Wang.
\newblock Bidirectional copy-paste for semi-supervised medical image
  segmentation.
\newblock In \emph{Proceedings of the IEEE/CVF Conference on Computer Vision
  and Pattern Recognition}, pages 11514--11524, 2023.

\bibitem[Basak and Yin(2023)]{basak2023pseudo}
Hritam Basak and Zhaozheng Yin.
\newblock Pseudo-label guided contrastive learning for semi-supervised medical
  image segmentation.
\newblock In \emph{Proceedings of the IEEE/CVF Conference on Computer Vision
  and Pattern Recognition}, pages 19786--19797, 2023.

\bibitem[Bernard et~al.(2018)Bernard, Lalande, Zotti, Cervenansky, Yang, Heng,
  Cetin, Lekadir, Camara, Ballester, et~al.]{bernard2018deep}
Olivier Bernard, Alain Lalande, Clement Zotti, Frederick Cervenansky, Xin Yang,
  Pheng-Ann Heng, Irem Cetin, Karim Lekadir, Oscar Camara, Miguel
  Angel~Gonzalez Ballester, et~al.
\newblock Deep learning techniques for automatic mri cardiac multi-structures
  segmentation and diagnosis: is the problem solved?
\newblock \emph{IEEE transactions on medical imaging}, 37\penalty0
  (11):\penalty0 2514--2525, 2018.

\bibitem[Berthelot et~al.(2019)Berthelot, Carlini, Goodfellow, Papernot,
  Oliver, and Raffel]{berthelot2019mixmatch}
David Berthelot, Nicholas Carlini, Ian Goodfellow, Nicolas Papernot, Avital
  Oliver, and Colin~A Raffel.
\newblock Mixmatch: A holistic approach to semi-supervised learning.
\newblock \emph{Advances in neural information processing systems}, 32, 2019.

\bibitem[Cao et~al.(2022)Cao, Wang, Chen, Jiang, Zhang, Tian, and
  Wang]{cao2022swin}
Hu Cao, Yueyue Wang, Joy Chen, Dongsheng Jiang, Xiaopeng Zhang, Qi Tian, and
  Manning Wang.
\newblock Swin-unet: Unet-like pure transformer for medical image segmentation.
\newblock In \emph{European conference on computer vision}, pages 205--218.
  Springer, 2022.

\bibitem[Chen et~al.(2020)Chen, Fan, Girshick, and He]{chen2020improved}
Xinlei Chen, Haoqi Fan, Ross Girshick, and Kaiming He.
\newblock Improved baselines with momentum contrastive learning.
\newblock \emph{arXiv preprint arXiv:2003.04297}, 2020.

\bibitem[Chen et~al.(2021)Chen, Yuan, Zeng, and Wang]{chen2021semi}
Xiaokang Chen, Yuhui Yuan, Gang Zeng, and Jingdong Wang.
\newblock Semi-supervised semantic segmentation with cross pseudo supervision.
\newblock In \emph{Proceedings of the IEEE/CVF Conference on Computer Vision
  and Pattern Recognition}, pages 2613--2622, 2021.

\bibitem[Du et~al.(2023)Du, Zhang, Liu, and Wang]{du2023coarse}
Jie Du, Xiaoci Zhang, Peng Liu, and Tianfu Wang.
\newblock Coarse-refined consistency learning using pixel-level features for
  semi-supervised medical image segmentation.
\newblock \emph{IEEE Journal of Biomedical and Health Informatics}, 2023.

\bibitem[He et~al.(2022)He, Dai, Zheng, Wu, Cao, Liu, Jiang, Yang, Huang, Si,
  et~al.]{he2022galaxy}
Wanwei He, Yinpei Dai, Yinhe Zheng, Yuchuan Wu, Zheng Cao, Dermot Liu, Peng
  Jiang, Min Yang, Fei Huang, Luo Si, et~al.
\newblock Galaxy: A generative pre-trained model for task-oriented dialog with
  semi-supervised learning and explicit policy injection.
\newblock In \emph{Proceedings of the AAAI conference on artificial
  intelligence}, number~10, pages 10749--10757, 2022.

\bibitem[Huang et~al.(2022)Huang, Chen, Xiong, Zhang, Chen, Sun, and
  Wu]{huang2022semi}
Wei Huang, Chang Chen, Zhiwei Xiong, Yueyi Zhang, Xuejin Chen, Xiaoyan Sun, and
  Feng Wu.
\newblock Semi-supervised neuron segmentation via reinforced consistency
  learning.
\newblock \emph{IEEE Transactions on Medical Imaging}, 41\penalty0
  (11):\penalty0 3016--3028, 2022.

\bibitem[Jiao et~al.(2022)Jiao, Zhang, Ding, Cai, and Zhang]{jiao2022learning}
Rushi Jiao, Yichi Zhang, Le Ding, Rong Cai, and Jicong Zhang.
\newblock Learning with limited annotations: a survey on deep semi-supervised
  learning for medical image segmentation.
\newblock \emph{arXiv preprint arXiv:2207.14191}, 2022.

\bibitem[Ke et~al.(2020)Ke, Qiu, Li, Yan, and Lau]{ke2020guided}
Zhanghan Ke, Di Qiu, Kaican Li, Qiong Yan, and Rynson~WH Lau.
\newblock Guided collaborative training for pixel-wise semi-supervised
  learning.
\newblock In \emph{Computer Vision--ECCV}, pages 429--445. Springer, 2020.

\bibitem[Kullback and Leibler(1951)]{kullback1951information}
Solomon Kullback and Richard~A Leibler.
\newblock On information and sufficiency.
\newblock \emph{The annals of mathematical statistics}, 22\penalty0
  (1):\penalty0 79--86, 1951.

\bibitem[Lai et~al.(2021)Lai, Tian, Jiang, Liu, Zhao, Wang, and
  Jia]{lai2021semi}
Xin Lai, Zhuotao Tian, Li Jiang, Shu Liu, Hengshuang Zhao, Liwei Wang, and
  Jiaya Jia.
\newblock Semi-supervised semantic segmentation with directional context-aware
  consistency.
\newblock In \emph{Proceedings of the IEEE/CVF Conference on Computer Vision
  and Pattern Recognition}, pages 1205--1214, 2021.

\bibitem[Li et~al.(2020)Li, Yu, Chen, Fu, Xing, and Heng]{li2020transformation}
Xiaomeng Li, Lequan Yu, Hao Chen, Chi-Wing Fu, Lei Xing, and Pheng-Ann Heng.
\newblock Transformation-consistent self-ensembling model for semisupervised
  medical image segmentation.
\newblock \emph{IEEE Transactions on Neural Networks and Learning Systems},
  32\penalty0 (2):\penalty0 523--534, 2020.

\bibitem[Li et~al.(2021)Li, Luo, Lin, Chen, and Heng]{li2021dual}
Yanwen Li, Luyang Luo, Huangjing Lin, Hao Chen, and Pheng-Ann Heng.
\newblock Dual-consistency semi-supervised learning with uncertainty
  quantification for covid-19 lesion segmentation from ct images.
\newblock In \emph{Medical Image Computing and Computer Assisted Intervention},
  pages 199--209. Springer, 2021.

\bibitem[Liang et~al.(2023)Liang, Wang, Miao, and Yang]{liang2023logic}
Chen Liang, Wenguan Wang, Jiaxu Miao, and Yi Yang.
\newblock Logic-induced diagnostic reasoning for semi-supervised semantic
  segmentation.
\newblock In \emph{Proceedings of the IEEE/CVF International Conference on
  Computer Vision}, pages 16197--16208, 2023.

\bibitem[Litjens et~al.(2014)Litjens, Toth, Van De~Ven, Hoeks, Kerkstra,
  Van~Ginneken, Vincent, Guillard, Birbeck, Zhang,
  et~al.]{litjens2014evaluation}
Geert Litjens, Robert Toth, Wendy Van De~Ven, Caroline Hoeks, Sjoerd Kerkstra,
  Bram Van~Ginneken, Graham Vincent, Gwenael Guillard, Neil Birbeck, Jindang
  Zhang, et~al.
\newblock Evaluation of prostate segmentation algorithms for mri: the promise12
  challenge.
\newblock \emph{Medical image analysis}, 18\penalty0 (2):\penalty0 359--373,
  2014.

\bibitem[Liu et~al.(2022)Liu, Desrosiers, and Zhou]{liu2022semi}
Jinhua Liu, Christian Desrosiers, and Yuanfeng Zhou.
\newblock Semi-supervised medical image segmentation using cross-model
  pseudo-supervision with shape awareness and local context constraints.
\newblock In \emph{International Conference on Medical Image Computing and
  Computer-Assisted Intervention}, pages 140--150. Springer, 2022.

\bibitem[Luo et~al.(2021{\natexlab{a}})Luo, Chen, Song, and Wang]{luo2021semi}
Xiangde Luo, Jieneng Chen, Tao Song, and Guotai Wang.
\newblock Semi-supervised medical image segmentation through dual-task
  consistency.
\newblock In \emph{Proceedings of the AAAI conference on artificial
  intelligence}, number~10, pages 8801--8809, 2021{\natexlab{a}}.

\bibitem[Luo et~al.(2021{\natexlab{b}})Luo, Liao, Chen, Song, Chen, Zhang,
  Chen, Wang, and Zhang]{luo2021efficient}
Xiangde Luo, Wenjun Liao, Jieneng Chen, Tao Song, Yinan Chen, Shichuan Zhang,
  Nianyong Chen, Guotai Wang, and Shaoting Zhang.
\newblock Efficient semi-supervised gross target volume of nasopharyngeal
  carcinoma segmentation via uncertainty rectified pyramid consistency.
\newblock In \emph{Medical Image Computing and Computer Assisted Intervention},
  pages 318--329. Springer, 2021{\natexlab{b}}.

\bibitem[Luo et~al.(2022)Luo, Hu, Song, Wang, and Zhang]{luo2022semi}
Xiangde Luo, Minhao Hu, Tao Song, Guotai Wang, and Shaoting Zhang.
\newblock Semi-supervised medical image segmentation via cross teaching between
  cnn and transformer.
\newblock In \emph{International Conference on Medical Imaging with Deep
  Learning}, pages 820--833. PMLR, 2022.

\bibitem[Milletari et~al.(2016)Milletari, Navab, and Ahmadi]{milletari2016v}
Fausto Milletari, Nassir Navab, and Seyed-Ahmad Ahmadi.
\newblock V-net: Fully convolutional neural networks for volumetric medical
  image segmentation.
\newblock In \emph{2016 fourth international conference on 3D vision (3DV)},
  pages 565--571. Ieee, 2016.

\bibitem[Ouali et~al.(2020)Ouali, Hudelot, and Tami]{ouali2020semi}
Yassine Ouali, C{\'e}line Hudelot, and Myriam Tami.
\newblock Semi-supervised semantic segmentation with cross-consistency
  training.
\newblock In \emph{Proceedings of the IEEE/CVF Conference on Computer Vision
  and Pattern Recognition}, pages 12674--12684, 2020.

\bibitem[Ren et~al.(2023)Ren, Li, Xu, Zhu, Wang, Liu, Chang, and
  Liang]{ren2023viewco}
Pengzhen Ren, Changlin Li, Hang Xu, Yi Zhu, Guangrun Wang, Jianzhuang Liu,
  Xiaojun Chang, and Xiaodan Liang.
\newblock Viewco: Discovering text-supervised segmentation masks via multi-view
  semantic consistency.
\newblock \emph{arXiv preprint arXiv:2302.10307}, 2023.

\bibitem[Ronneberger et~al.(2015)Ronneberger, Fischer, and
  Brox]{ronneberger2015u}
Olaf Ronneberger, Philipp Fischer, and Thomas Brox.
\newblock U-net: Convolutional networks for biomedical image segmentation.
\newblock In \emph{Medical Image Computing and Computer-Assisted Intervention},
  pages 234--241. Springer, 2015.

\bibitem[Sedai et~al.(2019)Sedai, Antony, Rai, Jones, Ishikawa, Schuman, Gadi,
  and Garnavi]{sedai2019uncertainty}
Suman Sedai, Bhavna Antony, Ravneet Rai, Katie Jones, Hiroshi Ishikawa, Joel
  Schuman, Wollstein Gadi, and Rahil Garnavi.
\newblock Uncertainty guided semi-supervised segmentation of retinal layers in
  oct images.
\newblock In \emph{Medical Image Computing and Computer Assisted Intervention},
  pages 282--290. Springer, 2019.

\bibitem[Shu et~al.(2022)Shu, Li, Xiao, Bi, and Li]{shu2022cross}
Yucheng Shu, Hengbo Li, Bin Xiao, Xiuli Bi, and Weisheng Li.
\newblock Cross-mix monitoring for medical image segmentation with limited
  supervision.
\newblock \emph{IEEE Transactions on Multimedia}, 2022.

\bibitem[Sohn et~al.(2020)Sohn, Berthelot, Carlini, Zhang, Zhang, Raffel,
  Cubuk, Kurakin, and Li]{sohn2020fixmatch}
Kihyuk Sohn, David Berthelot, Nicholas Carlini, Zizhao Zhang, Han Zhang,
  Colin~A Raffel, Ekin~Dogus Cubuk, Alexey Kurakin, and Chun-Liang Li.
\newblock Fixmatch: Simplifying semi-supervised learning with consistency and
  confidence.
\newblock \emph{Advances in neural information processing systems},
  33:\penalty0 596--608, 2020.

\bibitem[Tajbakhsh et~al.(2020)Tajbakhsh, Jeyaseelan, Li, Chiang, Wu, and
  Ding]{tajbakhsh2020embracing}
Nima Tajbakhsh, Laura Jeyaseelan, Qian Li, Jeffrey~N Chiang, Zhihao Wu, and
  Xiaowei Ding.
\newblock Embracing imperfect datasets: A review of deep learning solutions for
  medical image segmentation.
\newblock \emph{Medical Image Analysis}, 63:\penalty0 101693, 2020.

\bibitem[Tarvainen and Valpola(2017)]{tarvainen2017mean}
Antti Tarvainen and Harri Valpola.
\newblock Mean teachers are better role models: Weight-averaged consistency
  targets improve semi-supervised deep learning results.
\newblock \emph{Advances in neural information processing systems}, 30, 2017.

\bibitem[Verma et~al.(2022)Verma, Kawaguchi, Lamb, Kannala, Solin, Bengio, and
  Lopez-Paz]{verma2022interpolation}
Vikas Verma, Kenji Kawaguchi, Alex Lamb, Juho Kannala, Arno Solin, Yoshua
  Bengio, and David Lopez-Paz.
\newblock Interpolation consistency training for semi-supervised learning.
\newblock \emph{Neural Networks}, 145:\penalty0 90--106, 2022.

\bibitem[Wang et~al.(2019)Wang, Zhou, Shen, Park, Fishman, and
  Yuille]{wang2019abdominal}
Yan Wang, Yuyin Zhou, Wei Shen, Seyoun Park, Elliot~K Fishman, and Alan~L
  Yuille.
\newblock Abdominal multi-organ segmentation with organ-attention networks and
  statistical fusion.
\newblock \emph{Medical image analysis}, 55:\penalty0 88--102, 2019.

\bibitem[Wu et~al.(2022{\natexlab{a}})Wu, Wang, Song, Yang, and
  Qin]{wu2022cross}
Huisi Wu, Zhaoze Wang, Youyi Song, Lin Yang, and Jing Qin.
\newblock Cross-patch dense contrastive learning for semi-supervised
  segmentation of cellular nuclei in histopathologic images.
\newblock In \emph{Proceedings of the IEEE/CVF Conference on Computer Vision
  and Pattern Recognition}, pages 11666--11675, 2022{\natexlab{a}}.

\bibitem[Wu et~al.(2021)Wu, Xu, Ge, Cai, and Zhang]{wu2021semi}
Yicheng Wu, Minfeng Xu, Zongyuan Ge, Jianfei Cai, and Lei Zhang.
\newblock Semi-supervised left atrium segmentation with mutual consistency
  training.
\newblock In \emph{Medical Image Computing and Computer Assisted Intervention},
  pages 297--306. Springer, 2021.

\bibitem[Wu et~al.(2022{\natexlab{b}})Wu, Wu, Wu, Ge, and Cai]{wu2022exploring}
Yicheng Wu, Zhonghua Wu, Qianyi Wu, Zongyuan Ge, and Jianfei Cai.
\newblock Exploring smoothness and class-separation for semi-supervised medical
  image segmentation.
\newblock In \emph{International Conference on Medical Image Computing and
  Computer-Assisted Intervention}, pages 34--43. Springer, 2022{\natexlab{b}}.

\bibitem[Xu et~al.(2022)Xu, Zhou, Jin, Blumberg, Wilson, deGroot, Alexander,
  Oxtoby, and Jacob]{xu2022learning}
Mou-Cheng Xu, Yu-Kun Zhou, Chen Jin, Stefano~B Blumberg, Frederick~J Wilson,
  Marius deGroot, Daniel~C Alexander, Neil~P Oxtoby, and Joseph Jacob.
\newblock Learning morphological feature perturbations for calibrated
  semi-supervised segmentation.
\newblock In \emph{International Conference on Medical Imaging with Deep
  Learning}, pages 1413--1429. PMLR, 2022.

\bibitem[Yang et~al.(2022)Yang, Zhuo, Qi, Shi, and Gao]{yang2022st++}
Lihe Yang, Wei Zhuo, Lei Qi, Yinghuan Shi, and Yang Gao.
\newblock St++: Make self-training work better for semi-supervised semantic
  segmentation.
\newblock In \emph{Proceedings of the IEEE/CVF Conference on Computer Vision
  and Pattern Recognition}, pages 4268--4277, 2022.

\bibitem[Yang et~al.(2023{\natexlab{a}})Yang, Qi, Feng, Zhang, and
  Shi]{yang2023revisiting}
Lihe Yang, Lei Qi, Litong Feng, Wayne Zhang, and Yinghuan Shi.
\newblock Revisiting weak-to-strong consistency in semi-supervised semantic
  segmentation.
\newblock In \emph{Proceedings of the IEEE/CVF Conference on Computer Vision
  and Pattern Recognition}, pages 7236--7246, 2023{\natexlab{a}}.

\bibitem[Yang et~al.(2023{\natexlab{b}})Yang, Tian, Wan, Chen, Chen, and
  Chen]{yang2023semi}
Xiaosu Yang, Jiya Tian, Yaping Wan, Mingzhi Chen, Lingna Chen, and Junxi Chen.
\newblock Semi-supervised medical image segmentation via cross-guidance and
  feature-level consistency dual regularization schemes.
\newblock \emph{Medical Physics}, 2023{\natexlab{b}}.

\bibitem[Zhang et~al.(2023{\natexlab{a}})Zhang, Zhang, Tian, Lukasiewicz, and
  Xu]{zhang2023multi}
Shuo Zhang, Jiaojiao Zhang, Biao Tian, Thomas Lukasiewicz, and Zhenghua Xu.
\newblock Multi-modal contrastive mutual learning and pseudo-label re-learning
  for semi-supervised medical image segmentation.
\newblock \emph{Medical Image Analysis}, 83:\penalty0 102656,
  2023{\natexlab{a}}.

\bibitem[Zhang et~al.(2023{\natexlab{b}})Zhang, Jiao, Liao, Li, and
  Zhang]{zhang2023uncertainty}
Yichi Zhang, Rushi Jiao, Qingcheng Liao, Dongyang Li, and Jicong Zhang.
\newblock Uncertainty-guided mutual consistency learning for semi-supervised
  medical image segmentation.
\newblock \emph{Artificial Intelligence in Medicine}, 138:\penalty0 102476,
  2023{\natexlab{b}}.

\bibitem[Zhang et~al.(2023{\natexlab{c}})Zhang, Ran, Tian, Zhou, Li, Yang, and
  Jiao]{zhang2023self}
Zhenxi Zhang, Ran Ran, Chunna Tian, Heng Zhou, Xin Li, Fan Yang, and Zhicheng
  Jiao.
\newblock Self-aware and cross-sample prototypical learning for semi-supervised
  medical image segmentation.
\newblock \emph{arXiv preprint arXiv:2305.16214}, 2023{\natexlab{c}}.

\bibitem[Zhao et~al.(2023{\natexlab{a}})Zhao, Qi, Wang, Wang, Wu, Mao, and
  Zhang]{zhao2023rcps}
Xiangyu Zhao, Zengxin Qi, Sheng Wang, Qian Wang, Xuehai Wu, Ying Mao, and Lichi
  Zhang.
\newblock Rcps: Rectified contrastive pseudo supervision for semi-supervised
  medical image segmentation.
\newblock \emph{arXiv preprint arXiv:2301.05500}, 2023{\natexlab{a}}.

\bibitem[Zhao et~al.(2023{\natexlab{b}})Zhao, Yang, Long, Pi, Zhou, and
  Wang]{zhao2023augmentation}
Zhen Zhao, Lihe Yang, Sifan Long, Jimin Pi, Luping Zhou, and Jingdong Wang.
\newblock Augmentation matters: A simple-yet-effective approach to
  semi-supervised semantic segmentation.
\newblock In \emph{Proceedings of the IEEE/CVF Conference on Computer Vision
  and Pattern Recognition}, pages 11350--11359, 2023{\natexlab{b}}.

\bibitem[Zheng et~al.(2022{\natexlab{a}})Zheng, Xu, and Wei]{zheng2022double}
Ke Zheng, Junhai Xu, and Jianguo Wei.
\newblock Double noise mean teacher self-ensembling model for semi-supervised
  tumor segmentation.
\newblock In \emph{ICASSP 2022-2022 IEEE International Conference on Acoustics,
  Speech and Signal Processing (ICASSP)}, pages 1446--1450. IEEE,
  2022{\natexlab{a}}.

\bibitem[Zheng et~al.(2022{\natexlab{b}})Zheng, You, Huang, Wang, Qian, and
  Xu]{zheng2022simmatch}
Mingkai Zheng, Shan You, Lang Huang, Fei Wang, Chen Qian, and Chang Xu.
\newblock Simmatch: Semi-supervised learning with similarity matching.
\newblock In \emph{Proceedings of the IEEE/CVF Conference on Computer Vision
  and Pattern Recognition}, pages 14471--14481, 2022{\natexlab{b}}.

\bibitem[Zhou et~al.(2023)Zhou, Loy, and Liu]{zhou2023semi}
Kaiyang Zhou, Chen~Change Loy, and Ziwei Liu.
\newblock Semi-supervised domain generalization with stochastic stylematch.
\newblock \emph{International Journal of Computer Vision}, pages 1--11, 2023.

\bibitem[Zhou et~al.(2022)Zhou, Feng, Gu, Cheng, Lu, Shi, and
  Ma]{zhou2022uncertainty}
Qianyu Zhou, Zhengyang Feng, Qiqi Gu, Guangliang Cheng, Xuequan Lu, Jianping
  Shi, and Lizhuang Ma.
\newblock Uncertainty-aware consistency regularization for cross-domain
  semantic segmentation.
\newblock \emph{Computer Vision and Image Understanding}, 221:\penalty0 103448,
  2022.

\bibitem[Zou et~al.(2020)Zou, Zhang, Zhang, Li, Bian, Huang, and
  Pfister]{zou2020pseudoseg}
Yuliang Zou, Zizhao Zhang, Han Zhang, Chun-Liang Li, Xiao Bian, Jia-Bin Huang,
  and Tomas Pfister.
\newblock Pseudoseg: Designing pseudo labels for semantic segmentation.
\newblock \emph{arXiv preprint arXiv:2010.09713}, 2020.

\end{thebibliography}
